\documentclass[acmtog, nonacm]{acmart}

\AtBeginDocument{%
  }

\usepackage[utf8]{inputenc} 
\usepackage[T1]{fontenc}    
\usepackage{hyperref}       
\usepackage{url}            
\usepackage{booktabs}       
\usepackage{amsfonts}       
\usepackage{nicefrac}       
\usepackage{microtype}      
\usepackage{xcolor}         
\usepackage{graphicx}
\usepackage{algorithmic}
\usepackage{amsmath}
\usepackage{algorithm}
\usepackage{enumitem}
\usepackage{multirow}

\begin{document}

\title{Occlusion-Aware Physics-Semantic Keyframe Selection for Robust Video Editing}


\author{Lin Liu}
\authornote{Both authors contributed equally to this research.}
\affiliation{%
  \institution{Huawei}
  \country{China}}
\email{laulampaul@gmail.com}

\author{Zhihan Xiao}
\authornotemark[1] 
\affiliation{%
  \institution{Tsinghua University}
  \country{China}}
\email{xiaozh24@mails.tsinghua.edu.cn}

\author{Haohang Xu}
\affiliation{%
  \institution{East China Normal University}
  \country{China}}

\author{Rong Cong}
\affiliation{%
  \institution{Huawei}
  \country{China}}

\author{Zhibo Zhang}
\affiliation{%
  \institution{Huawei}
  \country{China}}

\author{Xiaopeng Zhang}
\affiliation{%
  \institution{Huawei}
  \country{China}}

\author{Qi Tian}
\authornote{Corresponding author.}
\affiliation{%
  \institution{Huawei}
  \country{China}}








\authorsaddresses{%
  Authors' Contact Information: 
  Lin Liu, Huawei, China, laulampaul@gmail.com; 
  Zhihan Xiao, Tsinghua University, China, xiaozh24@mails.tsinghua.edu.cn
}

\begin{teaserfigure}
  \centering
  \includegraphics[width=\linewidth]{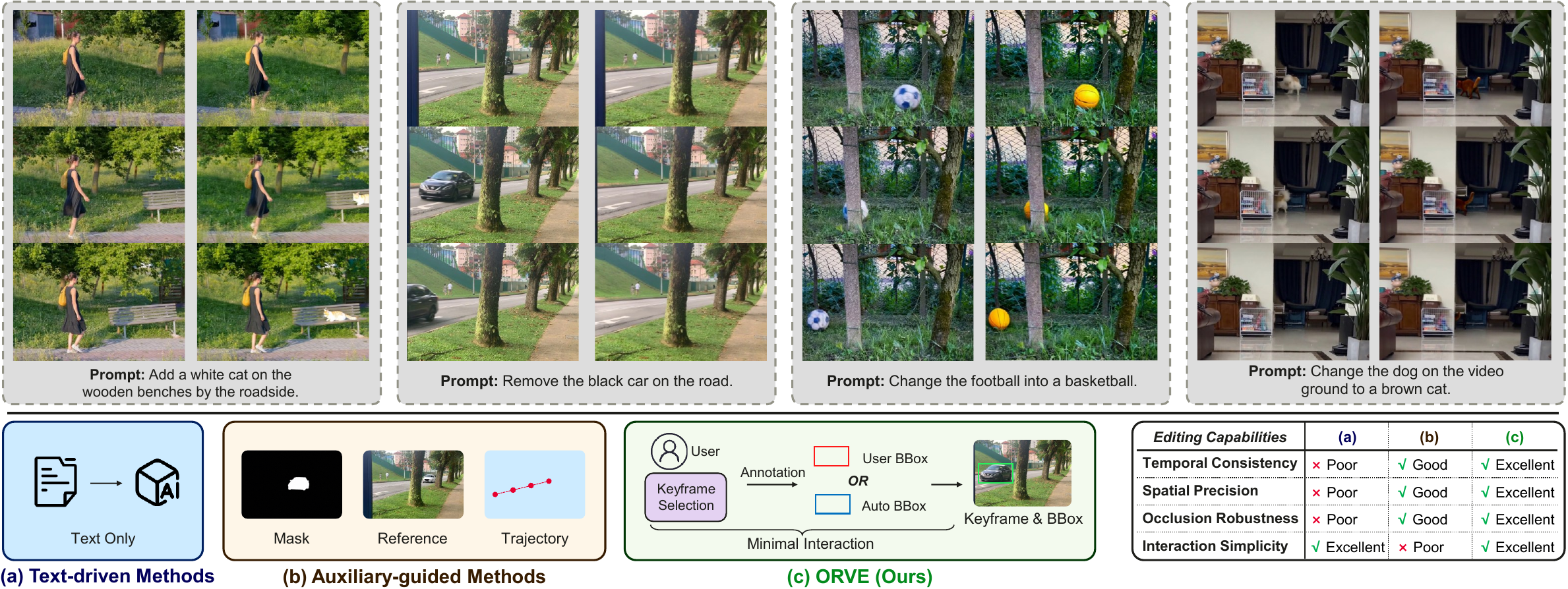}
\caption{
Comparison of video editing paradigms under occlusion. Unlike text-driven or manually guided methods, our approach identifies a reliable keyframe with minimal interaction, enabling robust and temporally consistent editing. 
}
  \label{fig:tease}
\end{teaserfigure}  

\begin{abstract}
  Video editing has recently achieved remarkable progress with diffusion-based generative models, enabling diverse object-level manipulations from natural language instructions. However, existing methods often struggle under occlusion, viewpoint changes, and fast object motion, where unreliable visual observations lead to inaccurate localization, temporal flickering, and inconsistent edits. In this work, we identify the absence of reliable visual anchors as a fundamental bottleneck in occlusion-robust video editing. To address this issue, we propose an occlusion-aware physics-semantic keyframe selection framework that automatically identifies an optimal anchor frame for downstream editing. Specifically, our method evaluates candidate frames from three complementary perspectives: structural completeness for avoiding truncated observations, cycle-consistent tracking stability for measuring physical reliability, and vision-language-based attribute visibility for ensuring semantic clarity. The selected keyframe is then propagated through bidirectional tracking to generate dense spatiotemporal masks, which are used as auxiliary supervision for a diffusion-based video editing backbone. By transforming occlusion handling from explicit reconstruction into reliable anchor selection, our framework enables precise and temporally consistent editing without requiring manual annotations. Extensive experiments on challenging video editing benchmarks demonstrate the effectiveness and high-quality performance of our method.
\end{abstract}





\maketitle

\section{Introduction}

Video editing has recently emerged as a fundamental task in generative modeling, enabling users to manipulate dynamic visual content through natural language instructions. With the rapid advancement of diffusion-based models, significant progress has been made in text-driven video editing, allowing diverse operations such as object insertion, removal, and attribute modification without requiring explicit supervision~\cite{bian2025videopainter,chen2025disco,wan2025unipaint,wang2023videocomposer,zhang2024avid,zi2025minimax}. Despite these advances, achieving temporally consistent and spatially precise editing remains a challenging problem, especially in complex real-world scenarios.

A major difficulty arises from \textbf{occlusion and viewpoint variation}, which frequently occur in dynamic scenes. When the target object is partially or fully occluded, or observed from unfavorable viewpoints, the visual evidence required for editing becomes ambiguous or unreliable. As a result, existing methods often suffer from artifacts such as temporal flickering, incorrect localization, or unintended modifications to background regions. Addressing occlusion remains particularly challenging, as it requires both identifying reliable visual evidence and maintaining consistent object correspondence across time.

Current video editing methods fall broadly into two paradigms. \textbf{Text-driven methods}~\cite{videocof,lucyedit, reco, icve} aim to perform editing solely based on language instructions, relying on implicit spatiotemporal reasoning within the generative model. While they offer high flexibility and minimal user effort, they often struggle with precise localization and temporal consistency under occlusion or complex motion, due to the lack of explicit structural constraints. On the other hand, \textbf{auxiliary-guided methods}~\cite{pisco,tu2025videoanydoor,Mtv-inpaint,phung2026traceobjectmotionediting} incorporate additional inputs such as masks, reference images, or trajectories to provide stronger spatial and temporal guidance. 
These methods improve controllability and stability, but typically require carefully prepared inputs and often become unreliable under severe occlusion, which limits their practicality in real-world applications. How to design guidance mechanisms that enable lightweight and intuitive user interaction, such as minimal user intervention for specifying target regions, in real-world scenarios remains underexplored.

In this work, we observe that the core bottleneck of occlusion-robust video editing lies in the absence of a \textbf{reliable visual anchor} that simultaneously satisfies structural completeness, temporal stability, and semantic clarity. Instead of directly addressing occlusion through reconstruction or heavy supervision, 
we propose to identify a visually reliable keyframe where the target object is fully observable, temporally stable, and semantically informative, and use it as an anchor for subsequent video editing.

To this end, we propose \textbf{ORVE}, an \textbf{O}cclusion-aware framework for \textbf{R}obust \textbf{V}ideo \textbf{E}diting based on physics-semantic keyframe selection. Our method evaluates candidate frames using three complementary criteria:  \emph{(i) structural completeness} to avoid truncated objects,  \emph{(ii) cycle-consistent tracking stability} to filter out frames affected by occlusion or abrupt motion, and \emph{ (iii) semantic attribute visibility} assessed by a vision-language model to ensure the edit-relevant attributes are observable. The selected keyframe is then propagated through bidirectional tracking to obtain a dense spatiotemporal mask, which serves as guidance for a diffusion-based video editing backbone. This design enables precise, temporally consistent editing without requiring manual annotations by default, while naturally supporting optional human intervention (e.g., user-specified bounding boxes).

Our approach provides a simple yet effective solution to occlusion-robust video editing by transforming the problem from explicit reconstruction to reliable anchor selection. Importantly, our framework unifies fully automatic and interactive editing paradigms, allowing flexible trade-offs between usability and controllability. Extensive experiments demonstrate that our method significantly improves editing quality under challenging scenarios involving occlusion, fast motion, and viewpoint changes. 

\textbf{Our contributions are summarized as follows:}
\begin{itemize}[leftmargin=1.2em]
    \item We introduce a novel Occlusion-aware Physics-Semantic Keyframe Selection strategy that jointly considers structural completeness, temporal stability, and semantic visibility.
    \item We propose a keyframe-conditioned video editing framework that leverages automatically generated spatiotemporal masks, largely reducing the need for manual annotations while supporting optional user guidance.
    \item We demonstrate that our method achieves superior performance in challenging scenarios with occlusion and complex dynamics, improving both spatial accuracy and temporal consistency. 
\end{itemize}
\section{Related Works}
\label{sec:related}
Existing video editing methods can be broadly categorized into two mainstream paradigms: (i) text-only approaches that rely on implicit consistency modeling ~\cite{ditto,insv2v,ku2024anyv2v,icve,lucyedit,VideoDirector,videocof,reco}, and (ii) auxiliary-guided methods that depend on explicit control signals such as masks, references, or user annotations ~\cite{chen2025disco,vace,occlusion,tu2025videoanydoor,zhang2025adaflowefficientlongvideo,point2insertvideoobjectinsertion,zhuang2025getinvideo},. While the former suffer from instability under complex dynamics, the latter require precise external inputs that are often expensive or impractical to obtain.
 
\subsection{Text-Driven Video Editing}
Recent progress in diffusion models has enabled video editing directly driven by textual instructions, without requiring additional control signals. In this paradigm, the model is expected to infer both \emph{where} and \emph{how} to edit solely from the input video and language description, offering a highly flexible and user-friendly interface.

Recent efforts have pushed this direction toward more powerful and general-purpose frameworks. For instance, Lucy-Edit~\cite{lucyedit} introduces a unified editing architecture that supports diverse operations such as object manipulation, attribute modification, and style transfer under pure text guidance. Ditto~\cite{ditto} constructs large-scale training data through an automated pipeline combining video filtering, image-based editing, and vision-language refinement, enabling robust instruction-following capabilities. ICVE~\cite{icve} leverages in-context generation with unpaired video clips to pretrain a unified editing model, improving generalization across diverse editing tasks. InsV2V~\cite{insv2v} improves text-video alignment by incorporating instruction-conditioned generation with additional consistency filtering, leading to more stable editing results across frames. VideoDirector~\cite{VideoDirector} further improves editing precision by leveraging text-to-video generation priors and carefully designed guidance strategies, enabling more accurate control over object-level modifications.


More recent works attempt to introduce finer control while remaining text-only. ReCo~\cite{reco} incorporates region-aware regularization into both latent and attention spaces, encouraging edits to concentrate on semantically relevant regions. VideoCoF~\cite{videocof} further introduces a structured reasoning paradigm, where edit-relevant tokens are first inferred through a chain-of-frames process before final video synthesis.

\subsection{Auxiliary-Guided Video Editing}

To overcome the limitations of purely text-driven approaches, a large body of work incorporates auxiliary signals to provide stronger spatial or temporal constraints. Depending on the form of guidance, these methods can be grouped into several categories.

\textbf{Reference-based Guidance.}
Reference-guided methods introduce exemplar images or objects to specify the desired appearance. Approaches such as GetInVideo~\cite{zhuang2025getinvideo} and VideoAnyDoor~\cite{tu2025videoanydoor} condition the generation process on one or more reference images, enabling precise control over object identity, texture, and style. By explicitly injecting appearance cues, these methods significantly reduce ambiguity in text descriptions. However, their performance depends on the availability and alignment of suitable reference inputs, which may not always be accessible in practical scenarios.

\textbf{Mask-based and Inpainting-based Guidance.}
Mask-based methods provide strong spatial control but struggle under occlusion and dynamic scenes. Early video inpainting methods (e.g., ProPainter~\cite{propainter}, MiniMax-Remover~\cite{zi2025minimax}) utilize binary masks to indicate regions for removal or modification, and reconstruct missing content by exploiting spatial-temporal redundancy. Building upon this paradigm, more advanced frameworks introduce unified architectures for both object removal and insertion. For example, MTV-Inpaint~\cite{Mtv-inpaint} employs a dual-branch attention mechanism to jointly handle foreground synthesis and background completion, while VideoPainter~\cite{bian2025videopainter} decouples context encoding and diffusion generation through a plug-and-play design, enabling scalable long-video editing.

\textbf{Spatial-Temporal Priors and Keyframe-based Guidance.}
A key limitation of conventional inpainting approaches is their sensitivity to occlusion and motion complexity. To address this, VOIN~\cite{occlusion} introduces a principled framework that explicitly models both \emph{shape completion} and \emph{flow completion}. Instead of directly hallucinating missing pixels, VOIN~\cite{occlusion} first reconstructs the complete object structure under occlusion and then estimates consistent motion fields to propagate appearance across frames. This decomposition significantly improves temporal coherence and robustness under severe occlusions, highlighting the importance of incorporating physical priors into video editing. Nevertheless, such methods are primarily designed for content recovery rather than semantic editing, and still rely on externally provided masks.

Recent works explore structured spatial-temporal priors to guide video editing. AdaFlow~\cite{zhang2025adaflowefficientlongvideo} exploits automatically selected keyframes to prune redundant tokens and improve efficiency for long video editing. More recently, TRACE~\cite{phung2026traceobjectmotionediting} introduces a trajectory-guided editing paradigm, where object motion is specified via first-frame trajectory annotations and propagated across time to control dynamic object behavior. Point2Insert~\cite{point2insertvideoobjectinsertion} pioneers a highly efficient point-based interaction paradigm, where a minimal set of user-provided points acts as a spatial anchor for object insertion. This drastically reduces manual annotation burdens while sustaining reasonable spatial accuracy. Building upon this user-friendly interactive concept, PISCO~\cite{pisco} further advances the methodology toward robust instance-level control. By introducing sparse yet highly structured guiding signals, PISCO facilitates seamless object insertion, culminating in substantially improved spatial precision and cross-frame temporal consistency.

These approaches highlight the importance of structured anchors, such as trajectories or keyframes, in stabilizing video editing. However, they typically rely on user-provided priors or predefined motion cues, and do not address how to automatically identify the most suitable anchor frame under complex conditions such as occlusion or viewpoint changes.

\section{Methodology}
\label{sec:method}
\begin{figure*}[t]
  \centering
  \includegraphics[width=\linewidth]{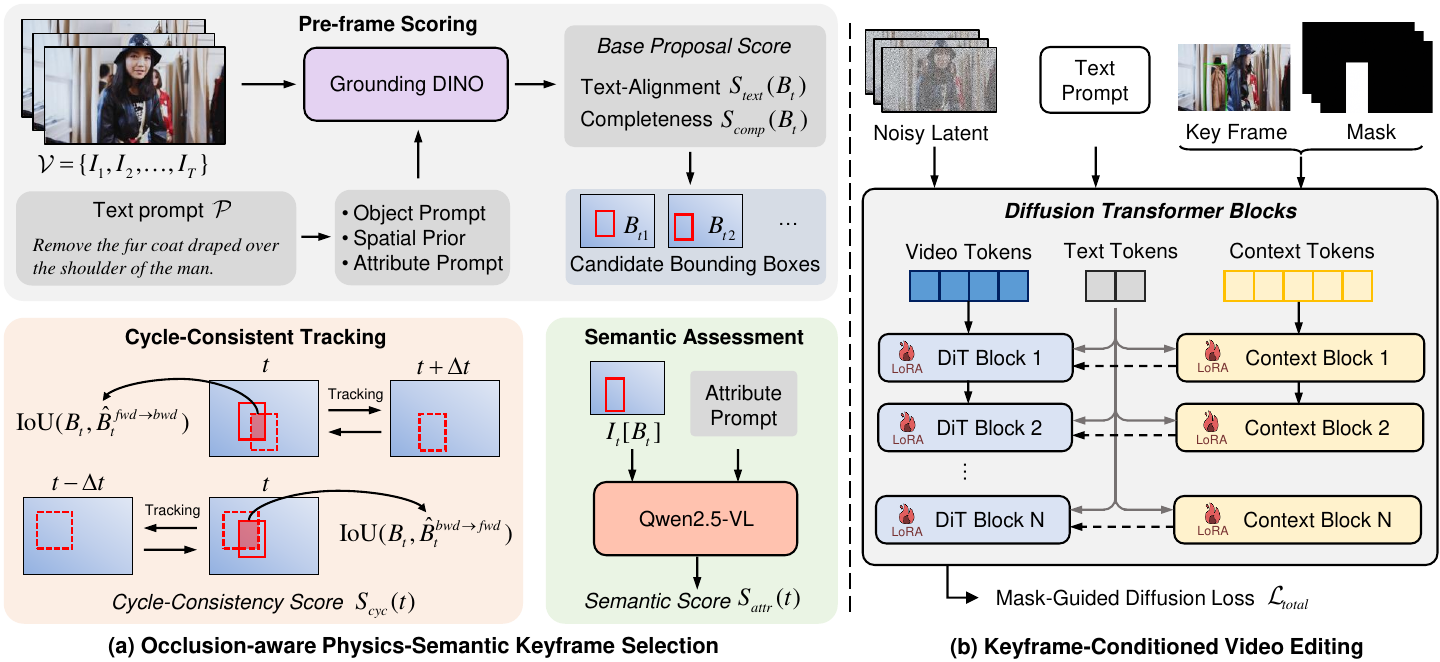}
\caption{
Overview of the proposed framework. Given an input video and a text prompt, an occlusion-aware physics-semantic keyframe selector identifies the optimal anchor frame, whose tracked masks are then injected into a 3D DiT to generate temporally consistent edited videos.
}
  \label{fig:pipeline}
\end{figure*}

\begin{figure*}[t]
  \centering
  \includegraphics[width=\linewidth]{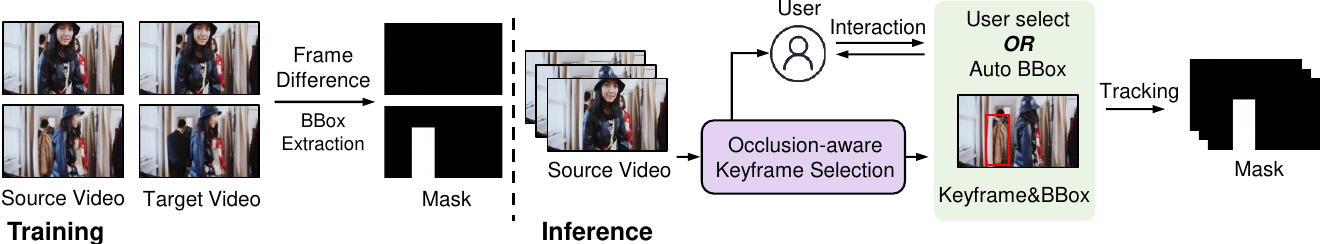}
\caption{Overview of the mask generation pipeline. During training, masks are generated from frame differences and bounding-box extraction; during inference, occlusion-aware keyframe selection with optional user interaction are used to produce masks.
}
  \label{fig:pipeline2}
\end{figure*}

Unlike existing approaches that rely on predefined or user-specified priors, we revisit video editing from the perspective of \textbf{anchor selection}. 
We argue that the key to robust video editing lies not in increasing supervision, but in identifying a reliable visual anchor that provides stable and unambiguous guidance for downstream generation. To this end, we propose a novel framework that decouples video editing into two sequential stages: \textbf{Occlusion-aware Physics-Semantic Keyframe Selection} (Sec.~\ref{subsec:keyframe}) and \textbf{Keyframe-Conditioned Video Editing} (Sec.~\ref{subsec:editing}) utilizing the ReCo~\cite{reco} diffusion backbone.  The overall pipeline is illustrated in Figure~\ref{fig:pipeline} and the mask generation pipeline is shown in Figure~\ref{fig:pipeline2}.


\subsection{Occlusion-aware Physics-Semantic Keyframe Selection}
\label{subsec:keyframe}
Given an input video sequence $\mathcal{V} = \{I_1, I_2, \dots, I_T\}$ and a specific user editing prompt $\mathcal{P}$, our goal is to select an optimal keyframe $I_{k^*}$ and obtain its corresponding object bounding box $B_{k^*}$, which can be either automatically detected or optionally specified by the user. A naive random selection or maximum-confidence selection is sub-optimal, as the target object might suffer from motion blur, severe occlusion, or unfavorable viewing angles that obscure the attributes to be edited. Therefore, we formulate the keyframe selection as an optimization problem maximizing a joint utility function incorporating structural completeness, temporal tracking stability, and semantic attribute visibility.

A high-quality keyframe for video editing must simultaneously satisfy three orthogonal properties:
\textit{(i) detectability}, ensuring that the target object is correctly localized;
\textit{(ii) physical stability}, ensuring that the object can be reliably tracked across time;
and \textit{(iii) semantic visibility}, ensuring that the prompt-relevant attribute is clearly observable.
These three factors are complementary and jointly sufficient for identifying an optimal visual anchor.

We first extract candidate bounding boxes for each frame $t \in \{1, \dots, T\}$ using an open-vocabulary detector GroundingDINO~\cite{grounding}. Optionally, our framework allows users to provide a bounding box on a selected frame, which serves as a more precise initialization. To handle ambiguous prompts (e.g., ``make the left car blue''), we introduce a spatial parser that disentangles $\mathcal{P}$ into an object prompt $\mathcal{P}_{obj}$ and a spatial prior $\mathcal{S}$. For a set of proposed boxes $\mathcal{B}_t$ in frame $I_t$, the text-alignment score is denoted as $S_{text}(B_t)$.

To prevent the model from selecting frames where the object is partially out-of-bounds (which inevitably leads to incomplete edits), we introduce a \textbf{Completeness Penalty}. For a bounding box $B_t = (x_1, y_1, x_2, y_2)$ in an image of size $W \times H$, the distance to the closest image border is $d_{min} = \min(x_1, y_1, W-x_2, H-y_2)$. The completeness score is defined as:
\begin{equation}
    S_{comp}(B_t) = \min\left(1, \max\left(0, \frac{d_{min}}{\tau \cdot \min(W, H)}\right)\right)
\end{equation}
where $\tau$ is a margin ratio. The base proposal score for a frame $t$ is thus $S_{base}(t) = S_{text}(B_t) \cdot S_{comp}(B_t)$. We filter the sequence to retain the top-$M$ candidate frames for further fine-grained screening.

\subsubsection{Temporal Stability via Cycle-Consistent Tracking}
An ideal keyframe should serve as a reliable anchor that can be robustly tracked across adjacent frames. To evaluate this physical stability, we propose a \textbf{Cycle-Consistent Tracking (CCT)} mechanism. Let $\Psi_{fwd}$ and $\Psi_{bwd}$ denote the forward and backward tracking operations. Specifically, we employ a temporal tracking algorithm as our probe. While our framework is agnostic to the underlying tracking architecture, we prioritize paradigms that are sensitive to structural deformations and computationally efficient. This component serves as an ideal ``critic'' for measuring temporal continuity by establishing local correspondences.

For a candidate box $B_t$ at frame $t$, we track it forward by $\Delta t$ frames and immediately track it backward to the original frame $t$, yielding $\hat{B}_{t}^{fwd \to bwd}$. A symmetric operation yields $\hat{B}_{t}^{bwd \to fwd}$. The cycle-consistency score $S_{cyc}(t)$ measures the Intersection over Union (IoU) between the original box and the cycle-tracked boxes:
\begin{equation}
    S_{cyc}(t) = \frac{1}{2} \Big( \text{IoU}(B_t, \hat{B}_{t}^{fwd \to bwd}) + \text{IoU}(B_t, \hat{B}_{t}^{bwd \to fwd}) \Big)
\end{equation}


A high cycle-consistency score $S_{cyc}(t)$ indicates that the forward-tracked box and its corresponding backward-recovered counterpart remain highly consistent under bidirectional motion estimation within the local temporal window $[t-\Delta t, t+\Delta t]$. Rather than providing a strict mathematical guarantee, this consistency serves as a reliable empirical indicator that the object is unlikely to experience severe occlusion, abrupt deformation, or tracking drift in this interval. Intuitively, if the object undergoes rapid motion, partial disappearance, or non-rigid transformation, the forward and backward tracking trajectories will diverge, resulting in reduced IoU and thus a lower $S_{cyc}(t)$, particularly under occlusion, where parts of the object become temporarily invisible. From this perspective, the cycle-consistency score can be interpreted as a proxy for local motion smoothness and structural stability, implicitly encoding a physical prior that favors frames where the object exhibits stable, predictable, and trackable dynamics. Consequently, maximizing $S_{cyc}(t)$ biases keyframe selection toward temporally coherent and physically reliable anchors for downstream editing.

\subsubsection{Semantic Clarity Assessment via Vision-Language Models}
\label{subsubsec:vlm}

While $S_{cyc}(t)$ ensures physical trackability, it does not guarantee that the \textit{editable attributes} are visually discernible. This is particularly important under occlusion, where the target attribute may be partially or fully invisible in certain frames. For instance, when the text-driven instruction $\mathcal{P}$ specifies ``change the logo on the shirt'', the keyframe must clearly expose the logo region rather than an occluded or non-informative viewpoint.

To address this limitation, we introduce a VLM-based semantic evaluator Qwen2.5-VL~\cite{qwen25vl} to assess attribute visibility in a zero-shot manner. We first parse the instruction $\mathcal{P}$ into a target attribute $A \in \{\text{color, material, part, shape, style}\}$ together with a spatial prior $\mathcal{S}$, which helps disambiguate multi-instance or spatially ambiguous queries.

Given a candidate bounding box $B_t$, we crop the corresponding image region $I_t[B_t]$ and prompt the VLM to evaluate the visibility of attribute $A$ on a normalized scale $[0,1]$. The resulting semantic score is defined as:
\begin{equation}
    S_{attr}(t) = \text{VLM}_{\theta}\Big(I_t[B_t], \text{Prompt}_{A}\Big)
\end{equation}

This semantic scoring mechanism complements detection-based confidence by focusing on \textit{attribute observability} rather than object presence. As a result, it ensures that the selected keyframe provides sufficient visual evidence for the intended edit operation, reducing ambiguity and mitigating failure cases such as occlusion-induced attribute hallucination or insufficient texture visibility.




\subsubsection{Optimal Keyframe Formulation}
The final optimal keyframe index $k^*$ is obtained by maximizing the joint utility function:
\begin{equation}
    k^* = \arg\max_{t \in \text{top-}M} \Big( \lambda_b S_{base}(t) + \lambda_c S_{cyc}(t) + \lambda_p S_{attr}(t) \Big)
\end{equation}
where $\lambda_b, \lambda_c, \lambda_p$ are hyperparameters balancing detection confidence, tracking stability, and semantic clarity, respectively. Once $I_{k^*}$ and $B_{k^*}$ are determined, we utilize the bidirectional tracker to propagate $B_{k^*}$ across the entire video, yielding a dense tube of masks $\mathcal{M} = \{M_1, \dots, M_T\}$.

\begin{algorithm}[t]
\caption{Physics- and Semantic-Driven Keyframe Selection}
\label{alg:keyframe}
\begin{algorithmic}[1]
\item[\textbf{Input:}] Video $\mathcal{V}=\{I_1, \dots, I_T\}$, Text prompt $\mathcal{P}$, Max delta $\Delta t$, Top-$M$
\item[\textbf{Output:}] Optimal keyframe index $k^*$, Tracked Masks $\mathcal{M}$ 

\STATE Parse $\mathcal{P}$ into object category $\mathcal{P}_{obj}$, spatial prior $\mathcal{S}$, and attribute $A$.
\FOR{$t = 1$ to $T$}
    \STATE $B_t, s_{text} \leftarrow \text{GroundingDINO}(I_t, \mathcal{P}_{obj}, \mathcal{S})$
    \STATE $S_{base}(t) \leftarrow s_{text} \times S_{comp}(B_t)$
\ENDFOR
\STATE $\mathcal{C} \leftarrow \text{Top-}M \text{ frames based on } S_{base}$
\FOR{$t \in \mathcal{C}$}
    \STATE Calculate $S_{cyc}(t)$ via Eq. 2 (Forward-backward tracking)
    \STATE Calculate $S_{attr}(t) \leftarrow \text{VLM}(I_t[B_t], A)$ via Eq. 3
    \STATE $S_{final}(t) \leftarrow \lambda_b S_{base}(t) + \lambda_c S_{cyc}(t) + \lambda_p S_{attr}(t)$
\ENDFOR
\STATE $k^* \leftarrow \arg\max_{t} S_{final}(t)$
\STATE $\mathcal{M} \leftarrow \text{BidirectionalTrack}(I_{k^*}, B_{k^*}, \mathcal{V})$
\RETURN $k^*, \mathcal{M}$
\end{algorithmic}
\end{algorithm}

\subsection{Keyframe-Conditioned Video Editing}
\label{subsec:editing}

Given the optimal keyframe $I_{k^*}$ and the corresponding spatiotemporal mask tube $\mathcal{M}$ obtained via bidirectional tracking, we perform video editing using a 3D DiT backbone. Unlike standard text-to-video generation, our setting requires preserving background content while selectively modifying foreground regions consistent with the instruction $\mathcal{P}$.

\textbf{Keyframe-Driven Spatiotemporal Conditioning.}
We first treat the selected keyframe $I_{k^*}$ as a visual anchor to establish consistent object semantics across time. Specifically, the object mask extracted from $I_{k^*}$ is propagated bidirectionally through the video sequence, yielding a dense mask tube $\mathcal{M} = \{M_1, \dots, M_T\}$. This propagation defines a consistent spatiotemporal correspondence between the reference frame and all target frames, which is further used to guide the editing process.

Rather than directly modifying pixel space, we inject the keyframe and mask information into the diffusion model as auxiliary conditioning signals. Concretely, the model takes as input the noisy latent $x_t$, prompt embedding $c$, and additional VACE-based~\cite{vace} structural conditions derived from the reference video and keyframe. These conditions are fused inside our 3D DiT backbone to provide spatial guidance for the denoising process.

\textbf{Mask-Guided Diffusion Supervision.}
Let $\epsilon_\theta(x_t, t, c)$ denote the noise prediction network, where $x_t$ is the noisy latent at timestep $t$, and $c$ represents the full conditioning set. To encourage precise editing while preserving background content, we adopt a region-aware diffusion objective defined as:
\begin{equation}
\mathcal{L}_{total} = \mathbb{E}_{x,\epsilon,t} \left[ \| \epsilon_\theta(x_t, t, c) - \epsilon \|_2^2 + \gamma \| (\epsilon_\theta(x_t, t, c) - \epsilon) \odot \mathcal{M} \|_2^2 \right],
\end{equation}
where $\odot$ denotes element-wise multiplication, $\mathcal{M}$ is the spatiotemporal mask broadcast to the latent space, and $\gamma$ controls the strength of foreground-focused supervision.

Unlike hard spatial masking approaches that directly modify input latents, our method uses $\mathcal{M}$ only as a supervisory signal in the diffusion objective. This design allows the model to retain global coherence while focusing representational capacity on editable regions. Combined with the keyframe-conditioned anchor $I_{k^*}$, the model learns a consistent mapping between text instructions and spatiotemporally aligned visual regions, enabling localized edits such as addition, removal, and replacement without degrading background content.

During inference, only the learned DiT backbone and conditioning pipeline are used, without requiring explicit mask enforcement.

\section{Experiments}
\label{sec:experiments}

\subsection{Experimental Setup}

\subsubsection{Implementation Details}
Our framework is built upon the ReCo~\cite{reco} video editing backbone and trained using the same large-scale instruction-driven video editing corpus released by ReCo, VideoCOF~\cite{videocof}, DITTO~\cite{ditto}, and OpenVE-3M~\cite{openve}. The dataset contains diverse editing scenarios including object addition, removal, and replacement, covering complex motions, viewpoint changes, and partial occlusions. To efficiently adapt the pretrained diffusion backbone to our keyframe-conditioned setting, we adopt LoRA for parameter-efficient finetuning.
Training is performed with an initial learning rate of $6 \times 10^{-5}$ using the AdamW optimizer. The model is trained for 8000 steps on 8 NVIDIA A800 GPUs with a total batch size of 8. For the keyframe selector, we set the completeness margin $\tau=0.05$, the temporal window $\Delta t=5$ for cycle-consistency tracking, and the top-candidate pool $M=5$. The balancing weights are empirically set to $\lambda_b = 0.5, \lambda_c = 0.3$, and $\lambda_p = 0.2$.

\subsubsection{Baseline Methods.}
We compare our approach with representative state-of-the-art video editing methods covering both open-source and commercial systems. For large-scale models ($\geq$14B), we include VACE~\cite{vace}, Ditto~\cite{ditto}, SAMA~\cite{sama}, and the closed-source Runway Aleph. For lightweight open-source models ($<14$B), we compare against ICVE~\cite{icve}, KiwiEdit~\cite{kiwiedit}, Lucy-Edit~\cite{lucyedit}, OpenVE-Edit~\cite{openve}, InsViE~\cite{wu2025insvie}, OmniVideo~\cite{yang2026omnivideo}, and our backbone model ReCo~\cite{reco}.

Among these baselines, text-only approaches rely solely on language instructions for implicit localization, whereas auxiliary-guided approaches exploit additional structural conditions such as masks or reference images. Since our method is built upon the same 1.3B ReCo backbone, the comparison with ReCo directly isolates the contribution of the proposed occlusion-aware keyframe selection and mask generation strategy.

\subsubsection{Benchmarks.}
We conduct evaluation on two complementary benchmarks. \emph{Open-VE Bench}~\cite{openve} is a recently proposed public benchmark for open-domain video editing, containing diverse real-world editing instructions across object replacement, removal, and insertion. To further evaluate robustness under severe occlusion and complex object motion, we additionally construct an \emph{Occlusion-Bench} dataset with manually curated video-prompt pairs.
The data for this evaluation set mostly comes from MOSE~\cite{MOSE}.
Compared with existing benchmarks, Occlusion-Bench contains significantly more challenging scenarios involving partial visibility, abrupt viewpoint changes, and long-term object interactions, providing a more targeted evaluation of occlusion-aware editing. For more information about Occlusion-Bench, please refer to the supplementary material.

\subsubsection{Evaluation Metrics.}
We assess editing performance from three complementary aspects: \emph{(i) Instruction fidelity } measures how accurately the generated video satisfies the editing command. Following EditVerse~\cite{ju2026editverseunifyingimagevideo}, we report both frame-level and video-level text alignment scores based on multimodal embedding similarity. \emph{(ii) Perceptual video quality} evaluates whether the edited video remains visually coherent after manipulation. Following the VBench protocol~\cite{vbench}, we adopt three criteria including subject consistency, background preservation, motion smoothness, and aesthetic quality, which are particularly informative for object-centric editing tasks. \emph{(iii) Semantic assessment} is conducted using the multimodal large language model MiniCPM-V2.6~\cite{minicpm}, which provides structured scores on prompt adherence and editing realism. Compared with embedding-based metrics, this evaluation better captures high-level semantic correctness and perceptual plausibility.

\begin{table}[h]
\centering
\caption{Performance Comparison of Different Methods on Open-VE Bench}
\label{tab:evaluation_results}
\resizebox{\columnwidth}{!}{
    \begin{tabular}{cllcccc}
    \toprule
    \textbf{Scale} & \textbf{Method} & \textbf{Size} & \textbf{Replace}$\uparrow$ & \textbf{Remove}$\uparrow$ & \textbf{Add}$\uparrow$ & \textbf{Average}$\uparrow$ \\
    \midrule
    \multirow{4}{*}{$\ge$ 14B} 
    & VACE                    & 14B         & 2.07 & 1.46 & 1.26 & 1.60 \\
    & DITTO                   & 14B         & 2.98 & 1.85 & 2.15 & 2.33 \\
    & SAMA                    & 14B         & \underline{3.93} & \underline{3.32} & \underline{2.54} & \underline{3.26} \\
    & Runway Aleph            & -           & \textbf{4.18} & \textbf{4.16} & \textbf{2.78} & \textbf{3.71} \\
    \midrule
    \multirow{8}{*}{$<$ 14B} 
    & InsViE                  & 2B          & 1.48 & 1.36 & 1.17 & 1.34 \\
    & OmniVideo               & 1.3B        & 1.14 & 1.14 & 1.36 & 1.21 \\
    & ICVE                    & 13B         & 2.57 & 2.51 & 1.97 & 2.35 \\
    & KiwiEdit-instruct-only & 5B+3B       & 2.57 & 2.71 & 2.25 & \underline{2.51} \\
    & Lucy-Edit               & 5B          & \underline{3.20} & 1.75 & \underline{2.30} & 2.42 \\
    & OpenVE-Edit             & 5B          & 2.98 & 1.85 & 2.15 & 2.33 \\
    & ReCo                    & 1.3B        & 2.35 & \underline{2.96} & 1.96 & 2.42 \\
    & Ours                    & 1.3B        & \textbf{3.87} & \textbf{2.98} & \textbf{2.62} & \textbf{3.16} \\
    \bottomrule
    \end{tabular}
}
\end{table}

\begin{table}[t]
\centering
\caption{
Comparison on the object replacement, removal, and addition task under the proposed Occlusion-Bench. 
We evaluate editing performance from three complementary perspectives, including instruction fidelity (denoted as \textbf{Inst. Fid.}), perceptual video quality, and semantic assessment. 
PF denotes prompt following, and EQ denotes edit quality. 
}
\label{tab:replacement}

\resizebox{\linewidth}{!}{
\begin{tabular}{c|c|cccc|c}
\toprule
\multirow{2}{*}{\textbf{Replacement}} &
\textbf{Inst. Fid.}$\uparrow$ &
\multicolumn{4}{c|}{\textbf{Perceptual Video Quality}$\uparrow$} &
\textbf{Semantic}$\uparrow$ \\

\cmidrule(lr){2-7}

& Frame / Video
& Subject
& Background
& Motion
& Aesthetic
& PF / EQ \\

\midrule

Kiwi-Edit
& 21.68 / 15.76 & 0.9093 & \underline{0.9293} & \textbf{0.9845} & \underline{0.4769} & 3.98 / 4.14 \\


SAMA
& \underline{23.19} / 16.48 & 0.9101 & 0.9243 & \underline{0.9835} & 0.4814 & \underline{4.35} / \textbf{4.46} \\

VideoCoF
& 22.98 / \underline{16.70} & 0.9090 & 0.9237 & 0.9814 & 0.4750 & 4.19 / 4.27 \\

LucyEdit
& 22.91 / 16.45 & 0.9146 & 0.9252 & 0.9801 & 0.4478 & 3.85 / 3.96 \\

ReCo
& 21.63 / 15.58 & \textbf{0.9216} & 0.9186 & 0.9832 & 0.3896 & 4.00 / 4.33 \\

\textbf{ORVE (Ours)}
& \textbf{23.45} / \textbf{16.98} & \underline{0.9211} & \textbf{0.9300} & \underline{0.9835} & \textbf{0.4849} & \textbf{4.41} / \underline{4.35} \\
\midrule \midrule 
\multirow{2}{*}{\textbf{Remove}} &
\textbf{Inst. Fid.}$\uparrow$ &
\multicolumn{4}{c|}{\textbf{Perceptual Video Quality}$\uparrow$} &
\textbf{Semantic}$\uparrow$ \\

\cmidrule(lr){2-7}

& Frame / Video
& Subject
& Background
& Motion
& Aesthetic
& PF / EQ \\
\midrule

Kiwi-Edit
& 21.46 / 12.33 & 0.9142 & \underline{0.9327} & 0.9837 & \underline{0.4585} & \underline{4.84} / 4.69 \\


SAMA
& \underline{22.23} / 12.59 & \textbf{0.9223} & 0.9310 & \textbf{0.9863} & 0.4576 & 4.70 / \textbf{4.95} \\

VideoCoF
& 22.03 / 13.19 & 0.9065 & 0.9233 & 0.9824 & 0.4581 & 4.79 / 4.69 \\

LucyEdit
& 22.18 / 13.99 & 0.9189 & 0.9128 & 0.9793 & 0.4476 & 4.81 / 4.63 \\

ReCo
& 22.13 / \underline{14.41} & \underline{0.9210}	& 0.9205	& \underline{0.9851}	& 0.3669 & 4.38 / 4.76 \\

\textbf{ORVE (Ours)}
& \textbf{23.17} / \textbf{14.87} & \textbf{0.9223} & \textbf{0.9330} & 0.9846 & \textbf{0.4628} & \textbf{4.95} / \underline{4.80} \\
\midrule \midrule 
\multirow{2}{*}{\textbf{Addition}} &
\textbf{Inst. Fid.}$\uparrow$ &
\multicolumn{4}{c|}{\textbf{Perceptual Video Quality}$\uparrow$} &
\textbf{Semantic}$\uparrow$ \\

\cmidrule(lr){2-7}

& Frame / Video
& Subject
& Background
& Motion
& Aesthetic
& PF / EQ \\

\midrule

Kiwi-Edit
& 20.90 / \underline{18.32} & 0.8870 & 0.9217 & \textbf{0.9852} & \underline{0.4617} & 3.24 / 4.30 \\


SAMA
& 21.02 / 17.37 & 0.8977 & 0.9185 & 0.9787 & \textbf{0.4703} & 3.33 / 4.09 \\

VideoCoF
& \underline{21.46} / 18.18 & 0.9041 & \underline{0.9243} & 0.9761 & 0.4528 & \textbf{3.75} / 4.31 \\

LucyEdit
& 20.58 / 16.57 & 0.8995 & 0.9211 & 0.9798 & 0.4477 & 3.03 / 3.28 \\

ReCo
& 20.89 / 18.21 & \textbf{0.9177} & 0.9187 & \underline{0.9800} & 0.3597 & \textbf{3.75} / \underline{4.58} \\

\textbf{ORVE (Ours)}
& \textbf{21.98} / \textbf{18.48} & \underline{0.9053} & \textbf{0.9326} & 0.9749
& 0.4578 & \underline{3.47} / \textbf{4.78} \\

\bottomrule
\end{tabular}
}
\end{table}

\begin{figure}[t]
  \centering
  \includegraphics[width=\linewidth]{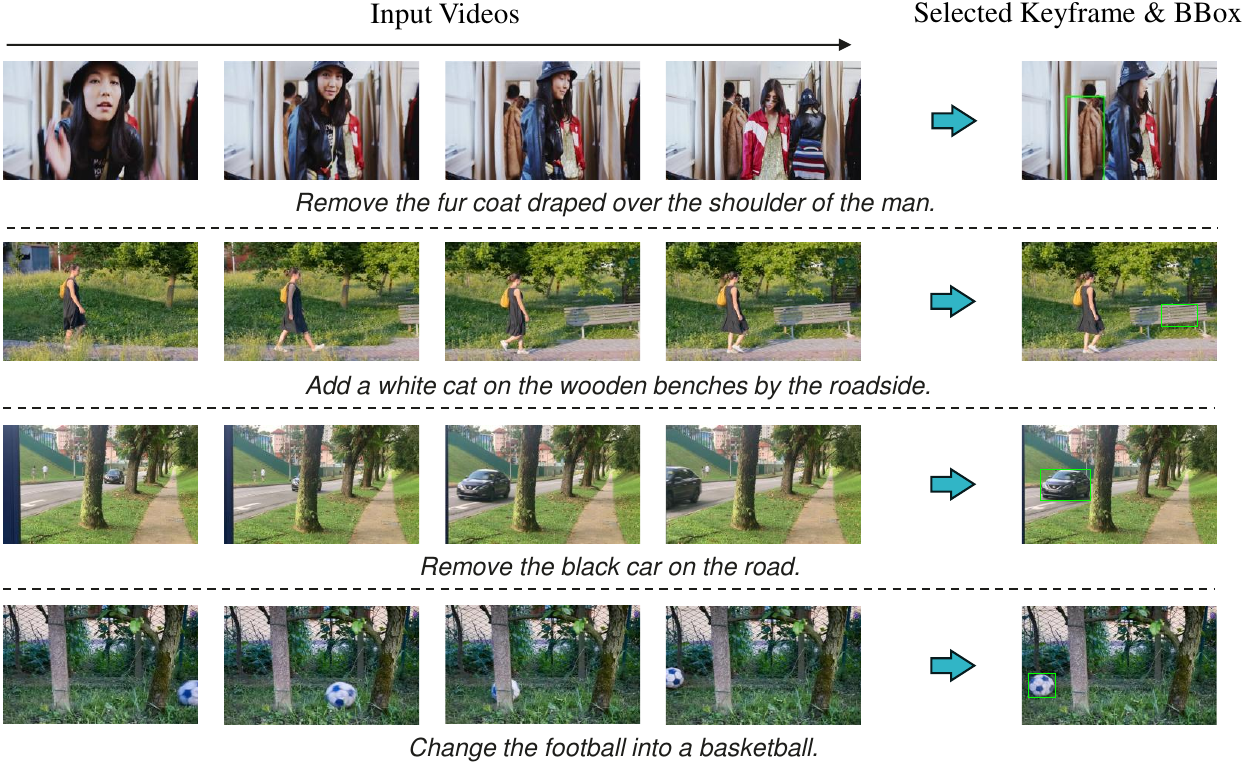}
\caption{
Visualization of the proposed keyframe selection strategy under occlusion scenarios. 
}
  \label{fig:keyframe_vis}
\end{figure}

\begin{figure}[t]
  \centering
  \includegraphics[width=\linewidth]{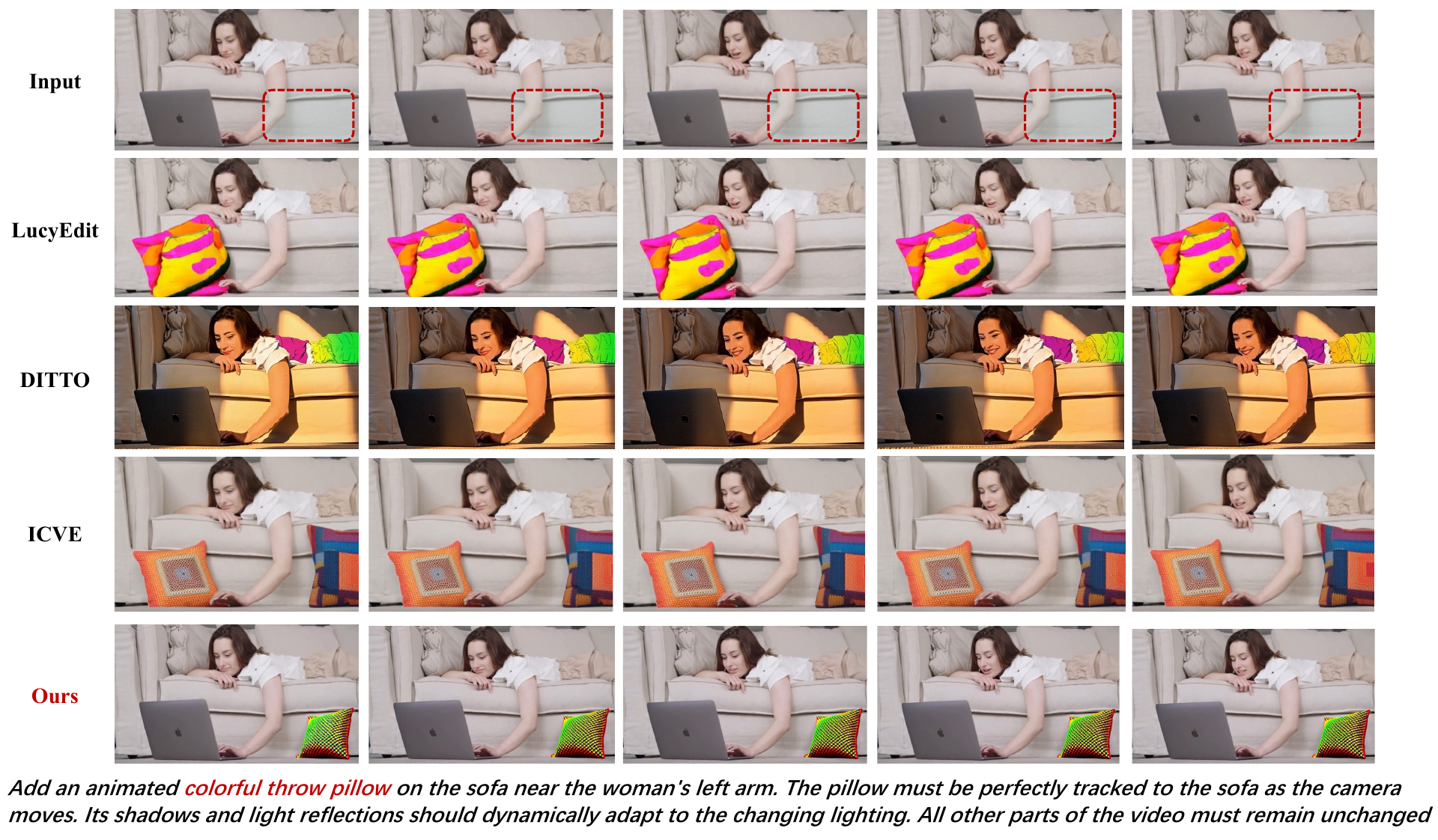}
    \includegraphics[width=\linewidth]{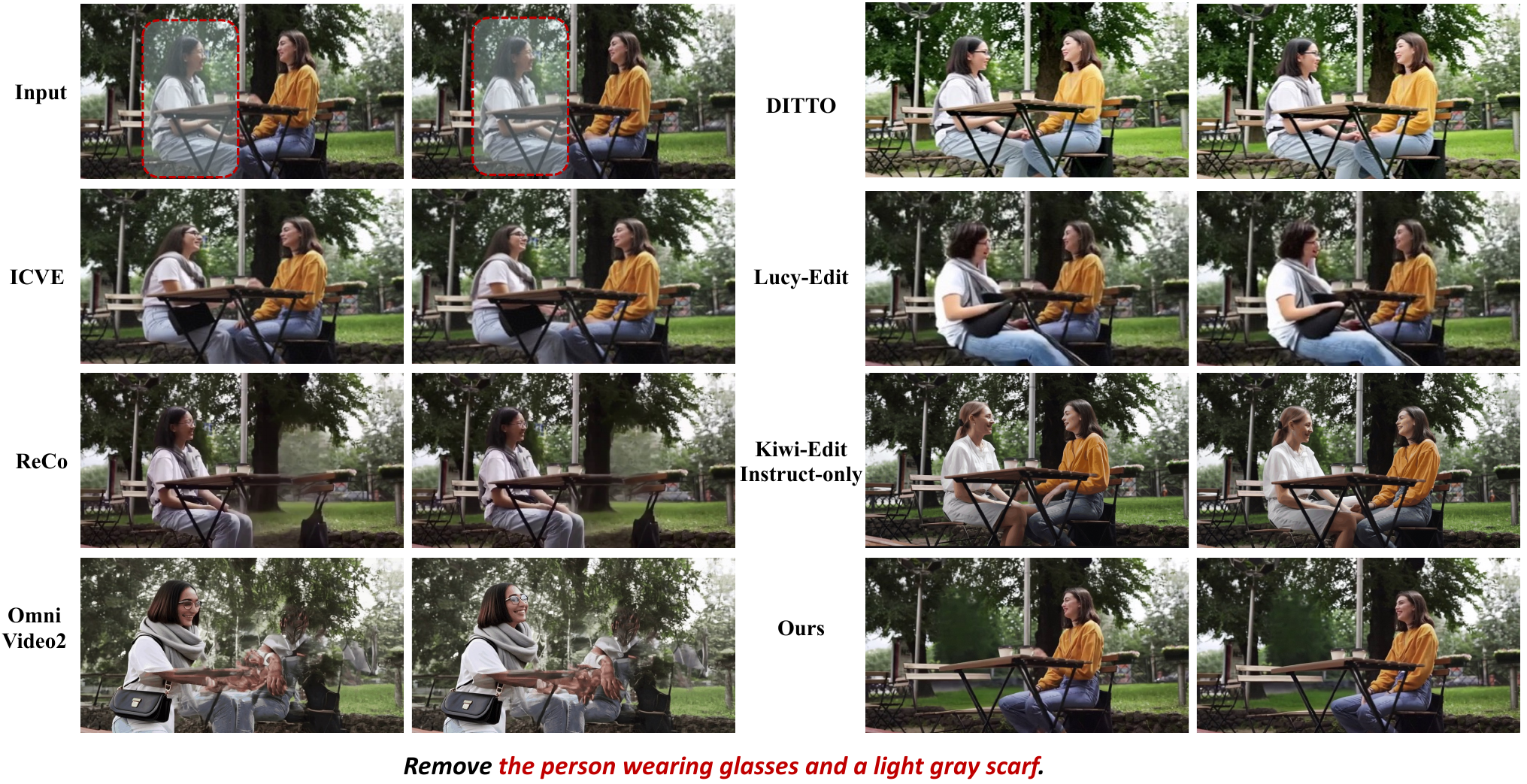}
      \includegraphics[width=\linewidth]{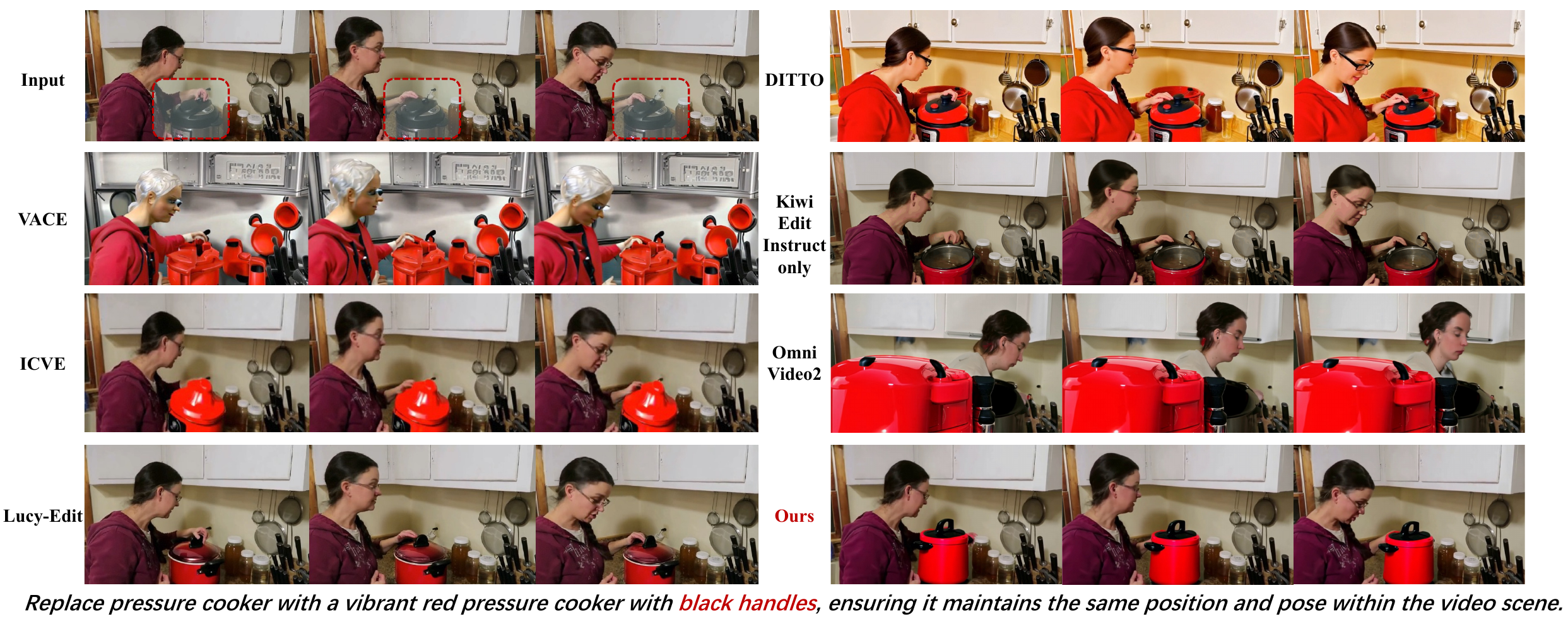}
  \vspace{-0.2cm}
\caption{Visual comparision between baseline methods on \textit{Addition}, \textit{Removal}, and \textit{Replacement} task.}
\vspace{-0.1cm}
  \label{fig:vis2}
\end{figure}



\subsection{Quantitative and Qualitative Comparisons}

\subsubsection{Results on Open-VE Bench.} 

Table~\ref{tab:evaluation_results} reports quantitative comparisons on Open-VE Bench. Our method achieves the best overall score among all open-source models, reaching an average VLM score of 3.16 while using only a 1.3B backbone.

Compared with our backbone model ReCo, our approach improves the average score from 2.42 to 3.16 (+30.6\%), demonstrating that the proposed occlusion-aware keyframe selection provides substantially more reliable editing conditions. Fig. ~\ref{fig:keyframe_vis} further visualizes the behavior of the proposed selector under challenging occlusion scenarios. 
Instead of relying on temporally central or randomly sampled frames, our method consistently selects frames where the target object remains structurally complete, semantically visible, and temporally stable. 
These selected anchors provide substantially more reliable conditions for subsequent diffusion-based editing.

\begin{figure*}[t]
  \centering
  \includegraphics[width=.85\linewidth]{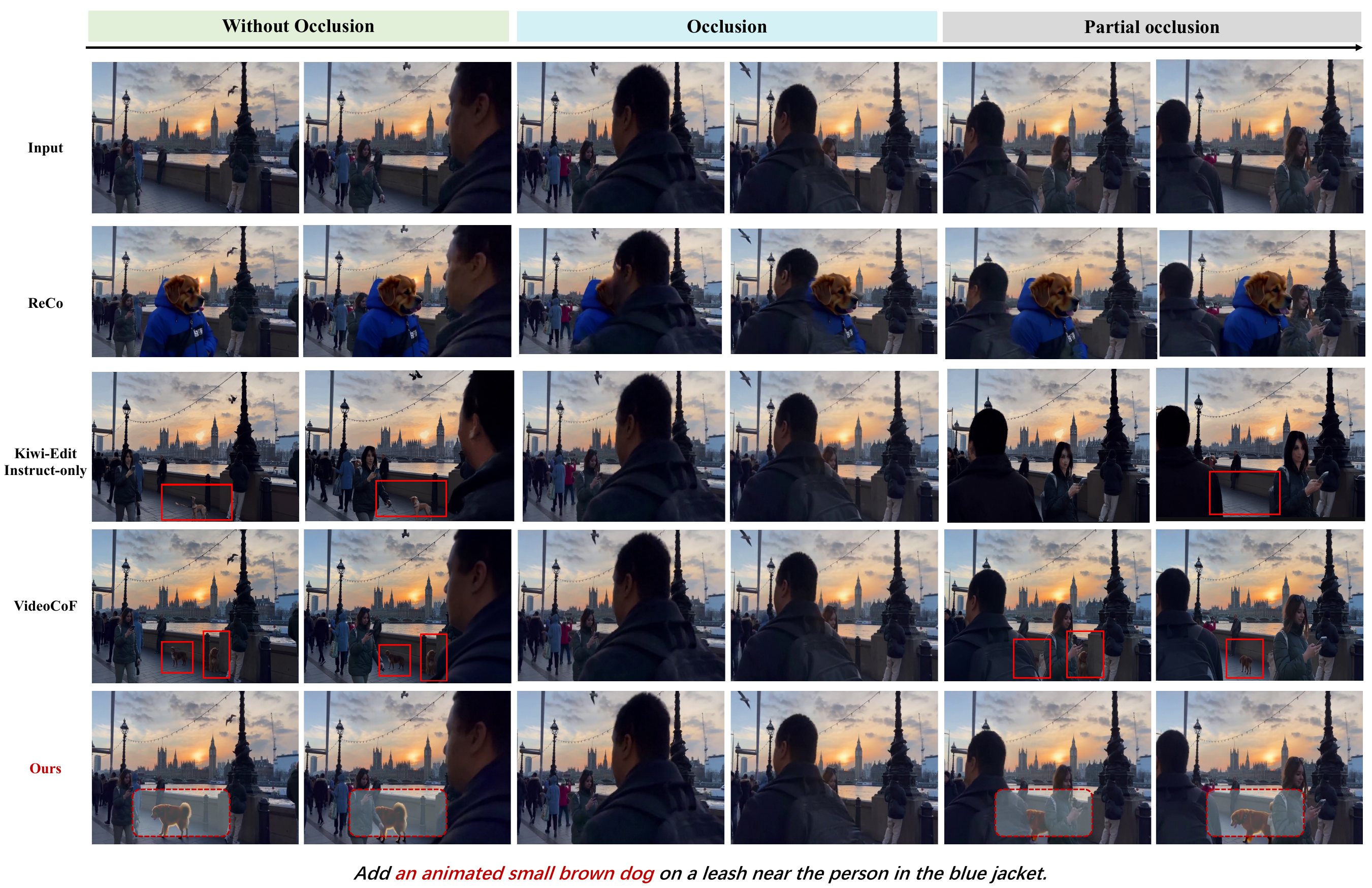}
  \vspace{-0.3cm}
\caption{Visualizations of video occlusion scenarios demonstrate that the proposed method achieves robust and consistently superior performance.}
\vspace{-0.15cm}
  \label{fig:vis5_occ}
\end{figure*}

\begin{figure*}[t]
  \centering
  \includegraphics[width=.85\linewidth]{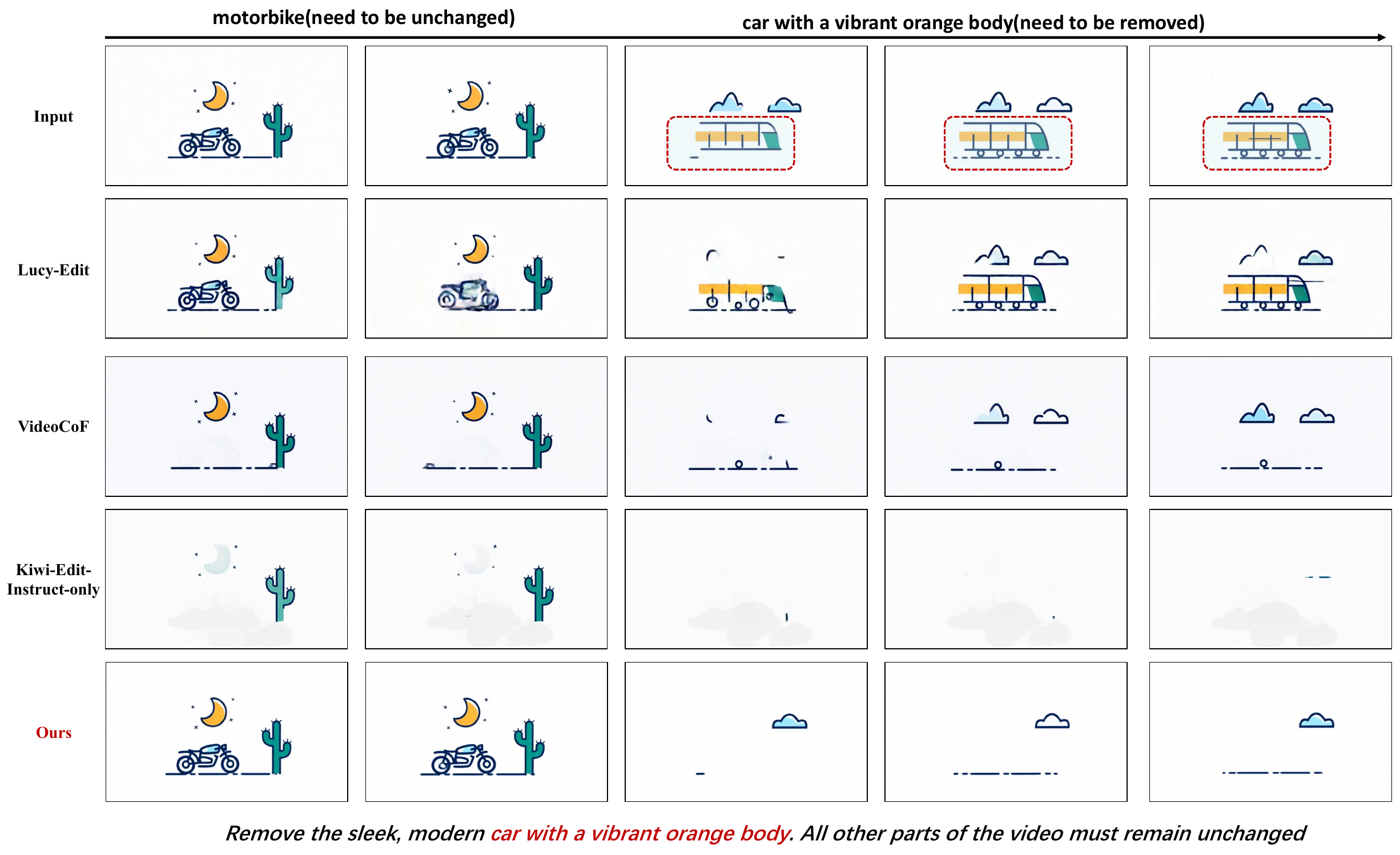}
\caption{The proposed method intelligently selects key frames, enabling temporal consistency and precise instruction following.}
  \label{fig:vis4_occ}
\end{figure*}

Notably, our method achieves the largest improvement on the \emph{replacement} task (2.35 $\rightarrow$ 3.87), which is consistent with our motivation that replacement requires accurate attribute visibility and precise object correspondence under viewpoint variations. Although some larger models contain substantially more parameters, our method still delivers competitive or superior performance, highlighting the effectiveness of reliable anchor selection over brute-force model scaling. Visual comparisons are presented in Figs.~\ref{fig:vis2}, where the proposed method shows a clear advantage over the baseline methods. Specifically, as shown in Figs.~\ref{fig:vis5_occ} and \ref{fig:vis4_occ}, it maintains robust and consistently superior performance even in complex occlusion scenarios.

\subsubsection{Results on Occlusion-Bench.}
\paragraph{Object Replacement.}
As shown in Table~\ref{tab:replacement}, object replacement is the most challenging setting, as it requires both precise spatiotemporal localization and fine-grained preservation of target attributes while simultaneously maintaining background consistency. Our proposed method consistently delivers strong performance across both perceptual and semantic metrics.


\emph{Object Removal.}
For object removal, our method achieves competitive performance. Compared with text-only baselines, the tracked spatiotemporal masks provide more reliable foreground localization, significantly reducing residual artifacts and background corruption.

\emph{Object Addition.}
For object insertion, our method consistently improves prompt-following accuracy over the backbone model. This suggests that the selected visual anchor provides stable spatial context, allowing the diffusion model to place new objects more naturally under complex scene dynamics. A visual comparison demonstrating object addition under occlusion, using a sample from Occlusion-Bench, is provided in the supplementary material.

\begin{table}[t]
\centering
\caption{
Ablation study of the proposed ORVE framework on OcclusionBench.
We analyze the impact of keyframe selection strategies, tracking modules, scoring weights, and mask-guided conditioning.
}
\label{tab:ablation}
\resizebox{\linewidth}{!}{
\begin{tabular}{c|c|cccc|c}
\toprule

\multirow{2}{*}{\textbf{Ablation Study}} &
\textbf{Inst. Fid.}$\uparrow$ &
\multicolumn{4}{c|}{\textbf{Perceptual Video Quality}$\uparrow$} &
\textbf{Semantic}$\uparrow$ \\

\cmidrule(lr){2-7}

& Frame / Video
& Subject
& Background
& Motion
& Aesthetic
& PF / EQ \\

\midrule

Random Keyframe
& 21.28 / 16.44 & 0.9108 & 0.9256 & 0.9818 & 0.4145  & 4.03 / 4.54 \\

Middle Frame
& 21.12 / 16.38 & 0.9133 & 0.9246 & 0.9809 & 0.3850  & 3.92 / 4.25 \\

w/o Semantic Score ($\lambda_p=0$)
& 21.73 / 15.82 & 0.9071 & 0.9213 & 0.9802 & 0.4096  &  \underline{4.09} / 4.19 \\

w/o Tracking Score ($\lambda_c=0$)
& 21.67 / 15.89 & 0.9109 & 0.9221 & \textbf{0.9844} & 0.3904 &  3.71 / 4.14 \\

w/o Mask Guidance
& 21.46 / 16.18 & 0.8955 & 0.9132 & 0.9775 & \underline{0.4679} &  3.96 / 4.23 \\

KCF Tracker
& \underline{22.34} / 16.08 & 0.9030 & 0.9251 & 0.9794 & 0.4599 & 3.83 / \underline{4.62} \\

Auto BBox
& 21.95 / \underline{16.71} & \textbf{0.9220} & \underline{0.9283} & \underline{0.9834} & 0.4047 & 3.76 / 4.27 \\

\textbf{FULL (SiamRPN Tracker)}
& \textbf{22.87} / \textbf{16.78} & \underline{0.9162} & \textbf{0.9318} & 0.9810 & \textbf{0.4685} &   \textbf{4.27} / \textbf{4.64} \\

\bottomrule
\end{tabular}
}
\end{table}

\subsection{Ablation and Parameter Study}

\subsubsection{Impact of Keyframe Selection.}
As shown in Table~\ref{tab:ablation}, replacing the proposed selector with random sampling or selecting the temporal center frame consistently degrades performance across all metrics. This confirms that identifying a reliable visual anchor is critical for robust editing under occlusion and viewpoint variation.

\subsubsection{Impact of Tracking Modules.}
While KCF~\cite{kcf} provides comparable performance, SiamRPN~\cite{sima} achieves better results with substantially lower computational cost, validating our choice of a lightweight correlation-based tracker.

\subsubsection{Impact of Physics-Semantic Scoring.}
Removing either the semantic visibility term ($\lambda_p=0$) or the cycle-consistency term ($\lambda_c=0$) leads to consistent performance drops. This demonstrates that temporal stability and semantic observability are both essential for selecting reliable keyframes.

\subsubsection{Impact of Mask-Guided Conditioning.}
Removing mask-guided supervision remains inferior to the full model. This verifies that the selected keyframe provides useful visual priors, while dense spatiotemporal masks further improve localization accuracy and temporal consistency.

\subsubsection{Impact of Bounding Box Initialization.}
We evaluate the impact of bounding box generation by comparing manual annotation against automatic detection. Using manually specified bounding boxes, which serves as our default configuration, achieves slightly better performance across evaluation metrics. This indicates that while automatic detection is highly competitive and effective without human intervention, a precise human-annotated spatial anchor can further refine the initialization of the tracking module, leading to marginally improved temporal consistency and downstream editing accuracy.

\section{Conclusion}

In this paper, we presented a novel framework to address the persistent challenges of occlusion and viewpoint variation in text-driven video editing. Recognizing that unreliable visual evidence is a primary cause of temporal flickering and spatial artifacts in existing generative models, we shifted the paradigm from explicit reconstruction to reliable anchor selection. Thus, we introduced an Occlusion-aware Physics-Semantic Keyframe Selection strategy that identifies the optimal frame by jointly evaluating structural completeness, cycle-consistent tracking stability, and semantic visibility. 

By propagating this automatically selected (or optionally user-specified) keyframe via bidirectional tracking, our framework generates a dense spatiotemporal mask that effectively guides the diffusion-based editing backbone. 
Extensive experiments demonstrate that our approach yields superior spatial precision and temporal consistency, particularly in challenging real-world scenarios characterized by heavy occlusions and complex dynamics.

\textbf{Limitations and Future Work.} Despite its effectiveness, our method is not without limitations. First, because our spatial guidance relies on bidirectional tracking, the editing quality is inherently bounded by the performance of the underlying tracking model. In cases of extreme motion blur or prolonged, complete occlusion, mask drift may still occur. Second, querying a vision-language model for semantic visibility evaluation introduces computational overhead during the preprocessing stage. 

For future work, we plan to explore end-to-end formulations where keyframe selection and temporal tracking are implicitly unified within the latent space of the diffusion model, further improving inference efficiency. Additionally, extending this reliable-anchor paradigm to handle multi-object interactions and 3D-aware, view-consistent video editing presents an exciting direction for advancing controllable video generation.


\bibliographystyle{ACM-Reference-Format}
\bibliography{sample-base}

\appendix
\section*{Appendix}
\section{Detail of \emph{Occlusion-Bench}}
Our Occlusion-Bench dataset consists of two parts. One part is carefully selected videos from the training set of MOSE (accounting for the majority). The other part is videos collected from the web where objects are occluded in specific frames (only a single-digit number of videos). The MOSE dataset is a video instance segmentation dataset, thus ensuring no overlap with the generation-based training data we use. After collecting the data, we manually screened and verified it, ultimately obtaining 33 videos with 81 frames each. These videos were manually assigned instructions for addition, deletion, and modification, forming the final evaluation set.

\begin{figure*}[htbp]
  \centering
  \includegraphics[width=.88\linewidth]{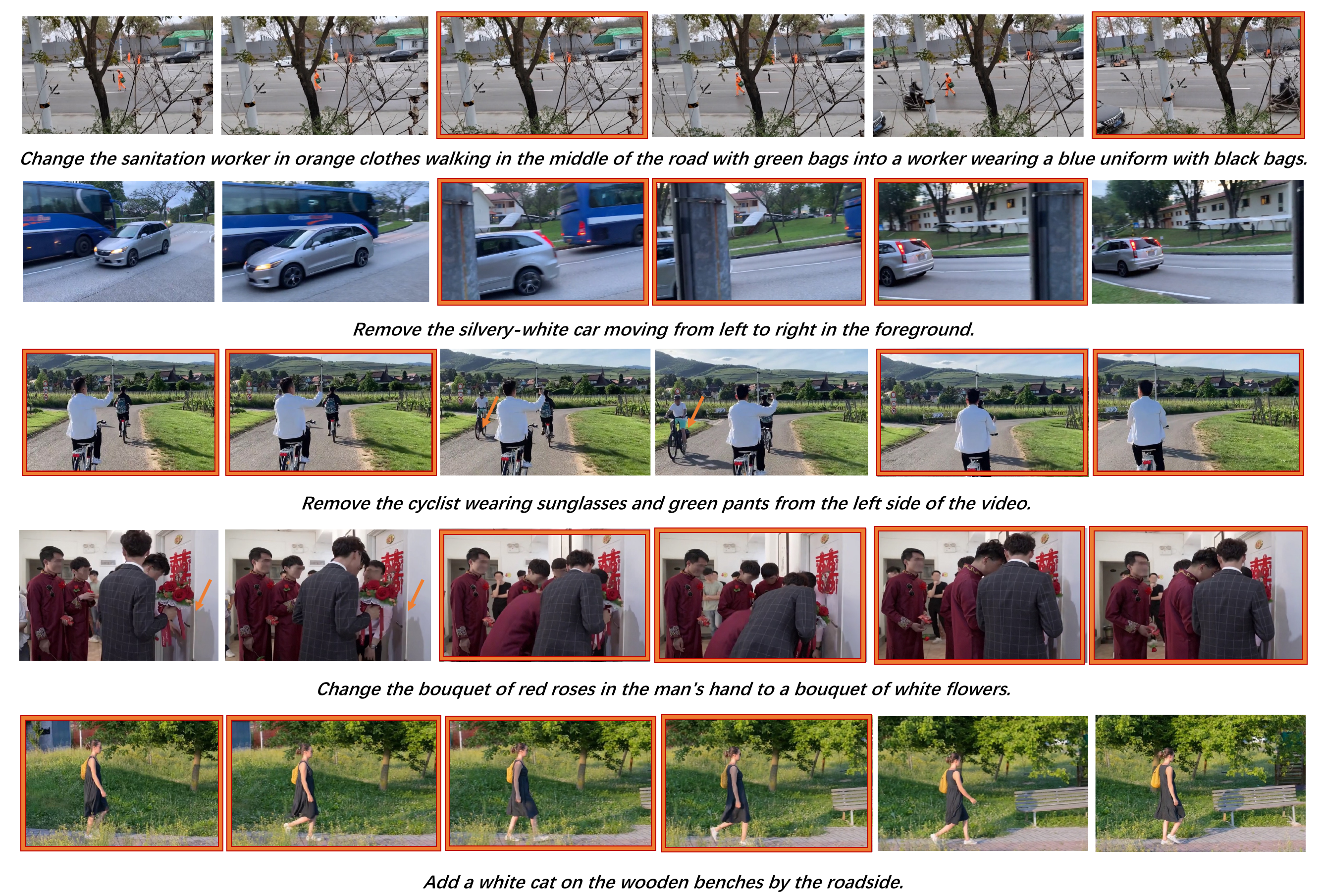}
\caption{More visualization examples of our proposed Occlusion-Bench. The frames in red box means that the object to be modified in the prompt is occluded.}
  \label{fig:vis5}
\end{figure*}

Fig.  \ref{fig:vis5} (If the frame where the object to be modified in the prompt is occluded, the shown frame will be marked with a \textcolor{red}{red box}.) below shows some visualizations of our dataset, indicating that our evaluation set has been carefully curated to ensure that certain objects appear during specific time periods and are occluded during other specific time periods, which better facilitates the evaluation of our task.

\section{More Visual Results}
\begin{figure*}[htbp]
  \centering
  \includegraphics[width=.88\linewidth]{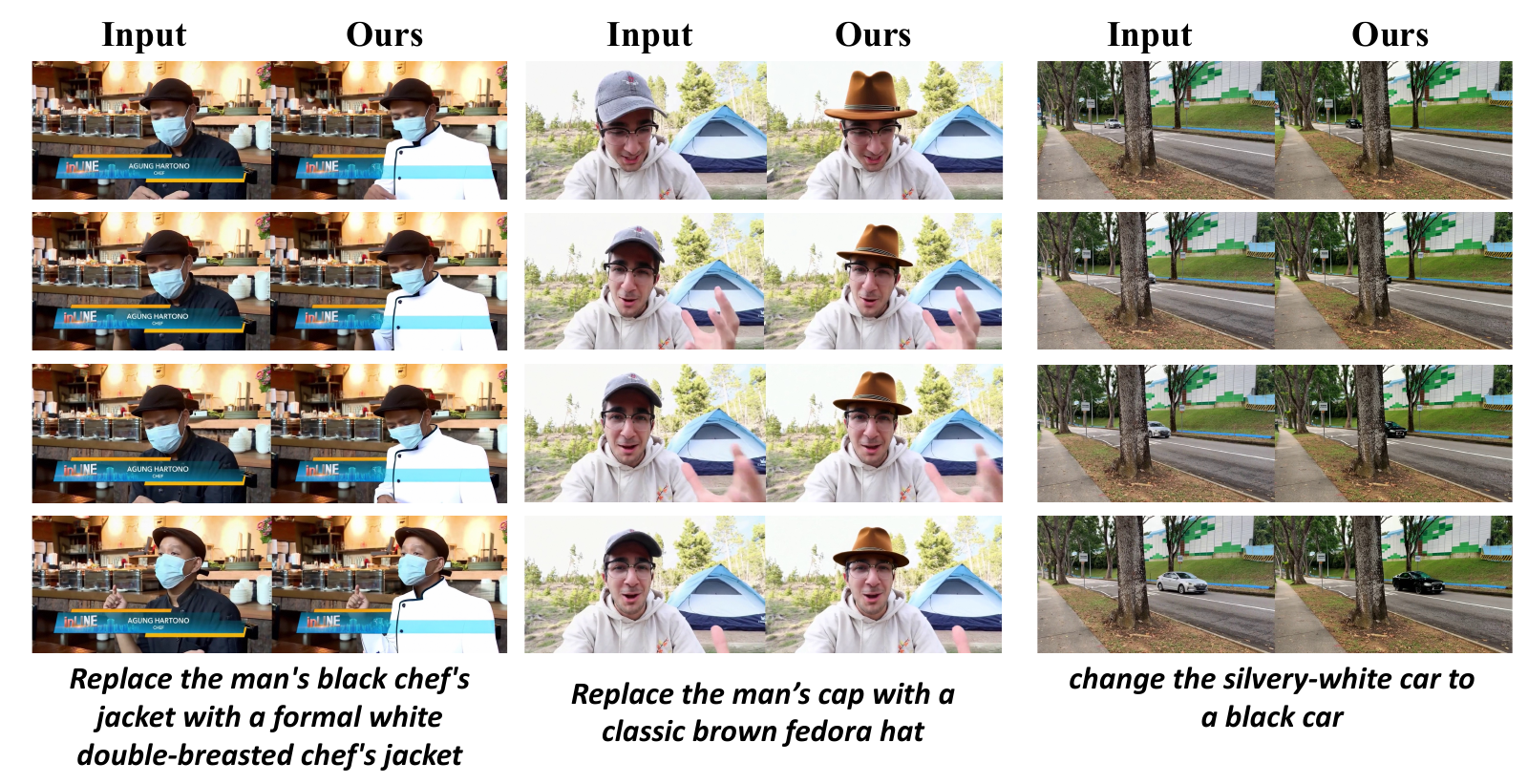}
\caption{More visualization results on \textit{replace} task (Samples are from Openve-Bench and Occlusion-Bench).}
  \label{fig:vis7}
\end{figure*}

\begin{figure*}[htbp]
  \centering
  \includegraphics[width=.88\linewidth]{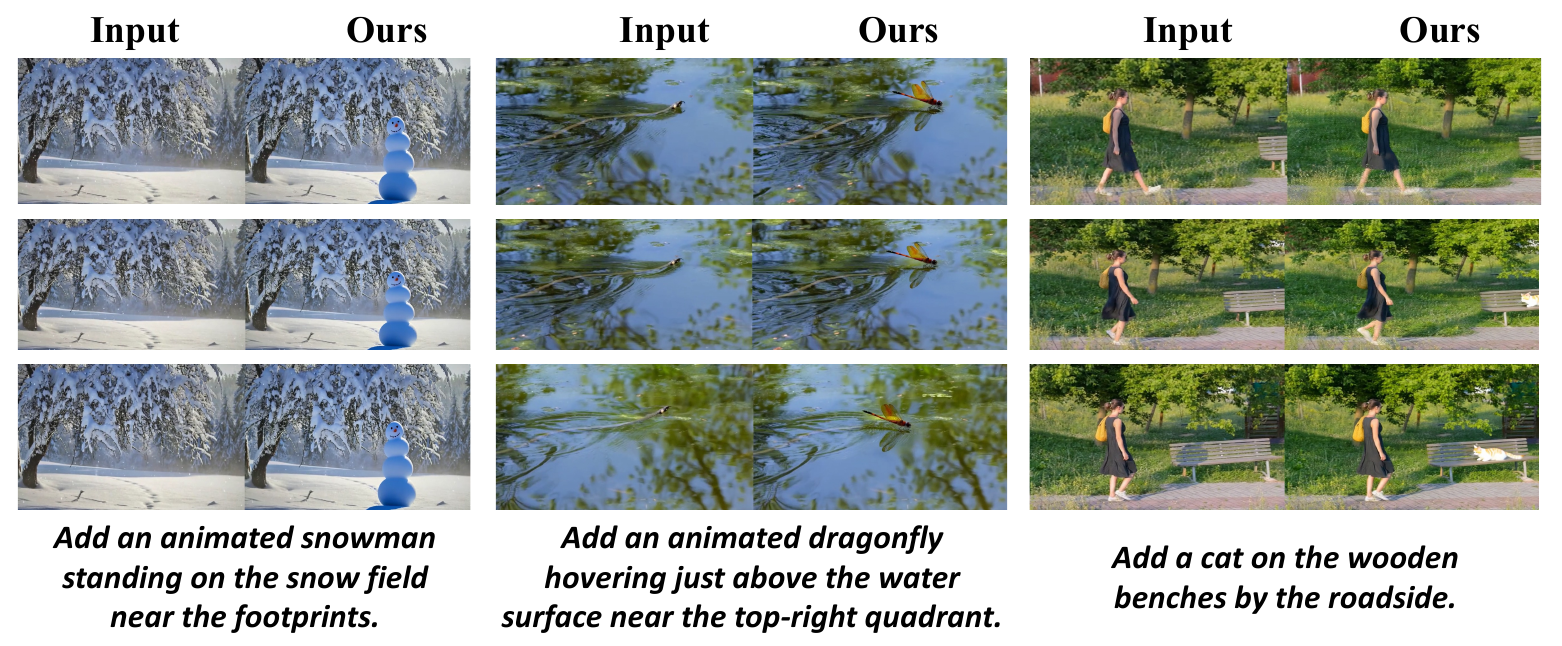}
\caption{More visualization results on \textit{add} task (Samples are from Openve-Bench and Occlusion-Bench).}
  \label{fig:vis8}
\end{figure*}

In the following figures (Fig.  \ref{fig:vis9}, Fig. \ref{fig:vis6}, Fig.  \ref{fig:vis7}, and Fig.  \ref{fig:vis8}), we present additional results on Reco-bench, Openve-bench, and Occlusion-bench. 

Fig.  \ref{fig:vis9} shows that SAMA incorrectly generated a wooden bench and a cat in the early frames. Kiwi-Edit missed the cat addition and unintentionally modified the bench. Meanwhile, LucyEdit mistakenly transformed the person into a cat.
Our method can successfully `locate' the wooden bench in later frames and add the cat to the correct position.


\begin{figure*}[htbp]
  \centering
  \includegraphics[width=.88\linewidth]{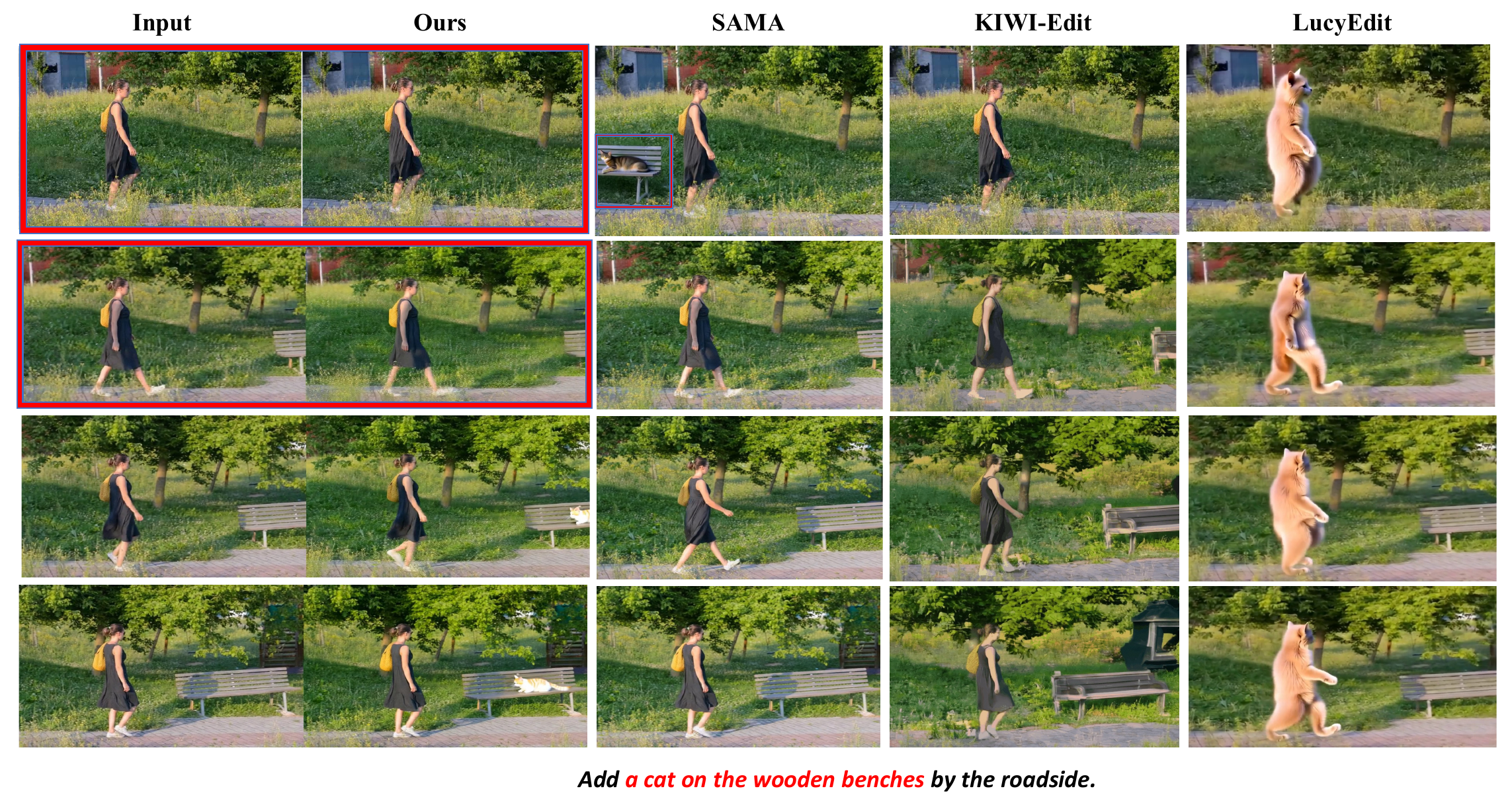}
\caption{Comparison of baseline methods on one \textit{add} example of Occlusion-Bench. SAMA incorrectly generated a wooden bench and a cat in the early frames. Kiwi-Edit missed the cat addition and unintentionally modified the bench. Meanwhile, LucyEdit mistakenly transformed the person into a cat.}
  \label{fig:vis9}
\end{figure*}

\begin{figure*}[htbp]
  \centering
  \includegraphics[width=.88\linewidth]{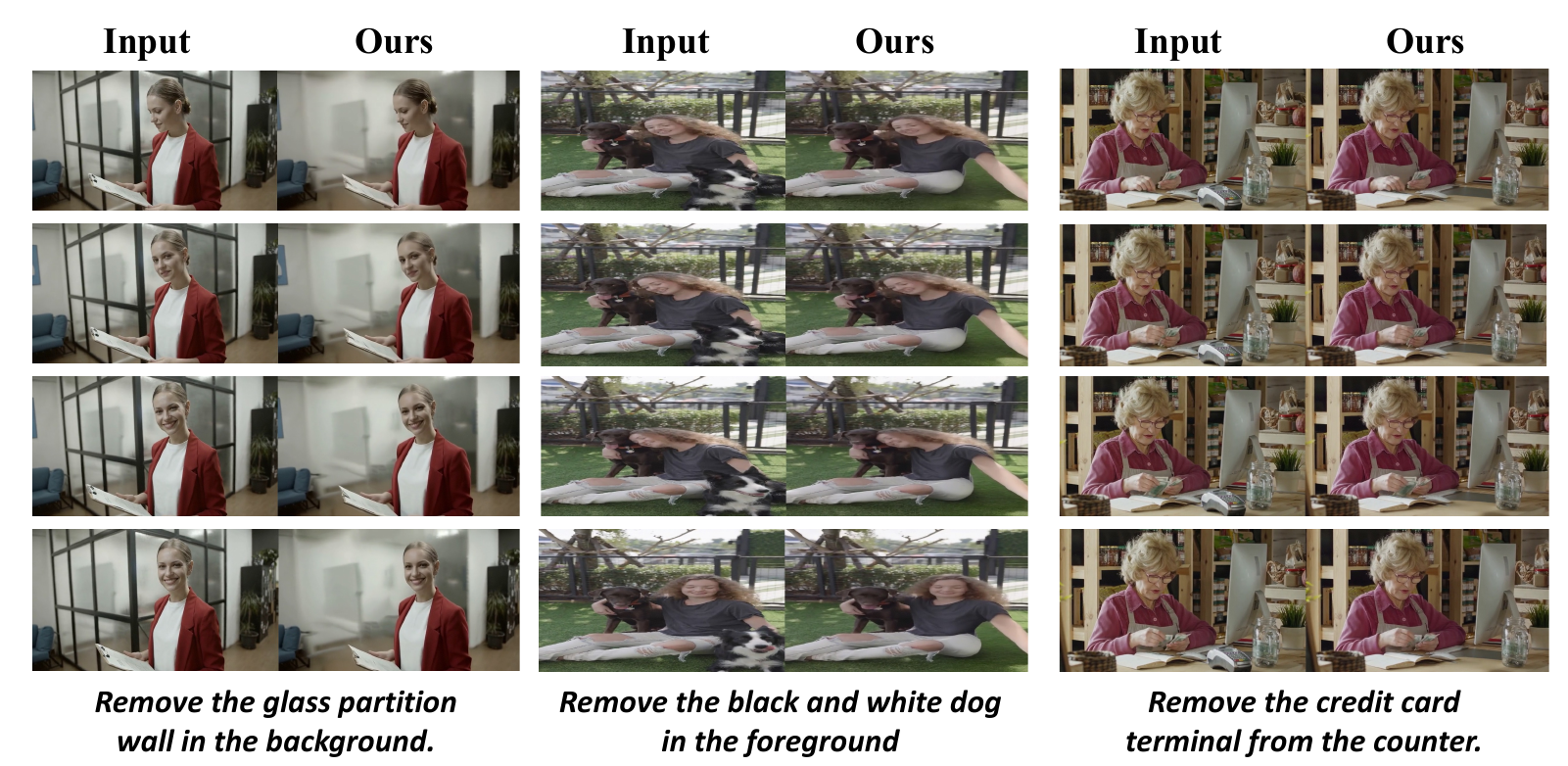}
\caption{More visualization results on \textit{remove} task of ReCo-Bench.}
  \label{fig:vis6}
\end{figure*}

\end{document}